\documentclass[letterpaper, 10 pt, conference]{ieeeconf}
\IEEEoverridecommandlockouts                              

\usepackage{times}
\usepackage{pstool}
\usepackage[space]{cite}
\usepackage{multicol}
\usepackage[bookmarks=true]{hyperref}

\usepackage{amsmath,amsfonts,amssymb} 
\usepackage[labelformat=simple]{subcaption}

\usepackage{pgfplots}
\pgfplotsset{compat=newest}
\usetikzlibrary{plotmarks}
\usetikzlibrary{arrows.meta}
\usepgfplotslibrary{patchplots}
\usepackage{grffile}
\usepackage{amsmath}
\newlength\fwidth
\usepackage{tikz}
\usepackage{tikz-3dplot}
\usetikzlibrary{patterns}

  \newcommand{\bmat}[1]{
  \begin{bmatrix}
      #1
  \end{bmatrix}
  }

\usepackage{xstring}

\begin{document}

\title{\LARGE \bf A Heteroscedastic Likelihood Model for Two-frame Optical Flow* 
}

\author{
Timothy Farnworth\authorrefmark{2}, %
Christopher Renton\authorrefmark{2}, %
Reuben Strydom\authorrefmark{3}, %
Adrian Wills\authorrefmark{2} and %
Tristan Perez\authorrefmark{3} 
\thanks{*This work was sponsored by Boeing Research \& Technology Australia.}
\thanks{\authorrefmark{2}
		Timothy Farnworth, Christopher Renton and Adrian Wills are with the Faculty of Engineering and Built Environment,
		The University of Newcastle,
		Callaghan, NSW 2308, Australia (e-mail:
        \href{mailto:timothy.farnworth@uon.edu.au}{\tt\small timothy.farnworth@uon.edu.au}; 
        \href{mailto:christopher.renton@newcastle.edu.au}{\tt\small christopher.renton@newcastle.edu.au}; 
        \href{mailto:adrian.wills@newcastle.edu.au}{\tt\small adrian.wills@newcastle.edu.au}
        )%
        }
\thanks{\authorrefmark{3} 
		Reuben Strydom and Tristan Perez are with Boeing Research \& Technology---Australia,
		St Lucia, QLD 4072, Australia (e-mail:
		\href{mailto:reuben.strydom@uqconnect.edu.au}{\tt\small reuben.strydom@uqconnect.edu.au}; 
        \href{mailto:tristan.perez@boeing.com}{\tt\small tristan.perez@boeing.com}
        )%
}
}

\maketitle

\begin{abstract}

Machine vision is an important sensing technology used in mobile robotic systems. 
Advancing the autonomy of such systems requires accurate characterisation of sensor uncertainty. 
Vision includes intrinsic uncertainty due to the camera sensor and extrinsic uncertainty due to environmental lighting and texture, which propagate through the image processing algorithms used to produce visual measurements. 
To faithfully characterise visual measurements, we must take into account these uncertainties.

In this paper, we propose a new class of likelihood functions that characterises the uncertainty of the error distribution of two-frame optical flow that enables a heteroscedastic dependence on texture.
We employ the proposed class to characterise the Farneb\"{a}ck and Lucas Kanade optical flow algorithms and achieve close agreement with their respective empirical error distributions over a wide range of texture in a simulated environment.
The utility of the proposed likelihood model is demonstrated in a visual odometry ego-motion study, which results in performance competitive with contemporary methods.
The development of an empirically congruent likelihood model advances the requisite tool-set for vision-based Bayesian inference and enables sensor data fusion with GPS, LiDAR and IMU to advance robust autonomous navigation.
\end{abstract}
\vspace{10pt}


\section{Introduction} 
\label{sec:introduction}

Navigation is a fundamental component of autonomous and semi-autonomous systems. 
Such systems are increasingly employed to perform tasks in maritime, land-based, aerial and space environments, which typically exploit GPS, IMU and/or vision measurements to ascertain their position and orientation (pose) within their environment.
Fusing the complementary sensor information is a challenging problem that relies on detailed knowledge of sensor characteristics, to be consolidated within a Bayesian inference framework.

Bayesian inference can be used to obtain the distribution $p(\mathbf{x}_k|\mathbf{y}_{1:k})$ of the system state $\mathbf{x}_k$ at time index $k$ conditioned on the sensor data $\mathbf{y}_{1:k} \equiv \mathbf{y}_1,\mathbf{y}_2,\dotsc,\mathbf{y}_k$ from the application of Bayes' rule and the Chapman-Kolmogorov equation~\cite{gregory2005bayesian},
\begin{align*}
    p(\mathbf{x}_k|\mathbf{y}_{1:k}) &= \frac{p(\mathbf{y}_k|\mathbf{x}_k)\,p(\mathbf{x}_k|\mathbf{y}_{1:k-1})}{\int_\mathcal{X} p(\mathbf{y}_k|\mathbf{x}_k)\,p(\mathbf{x}_k|\mathbf{y}_{1:k-1})\,\mathrm{d}\mathbf{x}_k}, \\
    p(\mathbf{x}_{k+1}|\mathbf{y}_{1:k}) &= \int_\mathcal{X} p(\mathbf{x}_{k+1}|\mathbf{x}_k)\,p(\mathbf{x}_k|\mathbf{y}_{1:k}) \,\mathrm{d}\mathbf{x}_k,
\end{align*}
where $p(\mathbf{y}_k|\mathbf{x}_k)$ is the sensor likelihood function, $p(\mathbf{x}_{k+1}|\mathbf{x}_k)$ is the state transition likelihood, and the recursion is initialised from some prior distribution $p(\mathbf{x}_1) = p(\mathbf{x}_1|\mathbf{y}_{-\infty:0})$ that takes into account all knowledge before time index $k=1$.
If the prior distribution, state transition likelihood and sensor likelihood functions are all Gaussian distributions, the solution to the Bayes filter is the celebrated Kalman filter~\cite{kalman1960new}.

As a result, it is convenient to assume GPS and IMU measurements exhibit Gaussian noise characteristics.
The fusion of these two complementary sensors provides pose and pose rate information.
With recent GPS spoofing attacks recorded in China and the Black Sea~\cite{c4ads2017:spoof}, GPS cannot be na\"{i}vely relied upon for autonomous system navigation. 
Therefore, it is desirable to augment vision into the sensor suite to cross-validate GPS data to enable robust, trusted autonomy. 
This requires a well characterised likelihood model for vision measurements.

Vision systems are unique in that measurement uncertainty is characterised by extrinsic environmental texture and lighting, much more so than the intrinsic noise of the camera sensor itself~\cite{wannenwetsch2017probflow,kendall2017uncertainties}. 
Vision systems typically compress the high-dimensional image data into a set of low-dimensional salient features in the scene, which may then treated as measurements. 
Due to the compression from raw image data to measured features, the distribution of measurement errors depends not only on the camera, lighting and environment, but also the image processing algorithms employed.

Optical flow is a vector field that describes the temporal evolution of an image and can be calculated from the motion of pixels across frame pairs.
This may be achieved using sparse methods, which typically look at the motion of image patches \cite{bouguet2001pyramidal}, or dense methods, which infer a per-pixel flow field across the image.
Dense methods use either variational techniques \cite{horn1981determining}, enforce smooth motion models across image patches \cite{farneback2003two}, or use machine learning techniques to infer flow based on relational structures observed during training \cite{dosovitskiy2015flownet,Sun2018PWC-Net}.

Optical flow can be used to determine the motion of a camera (ego-motion) over a sequence of images, since the flow field is the projection of relative motion between the camera and the environment. 
Because image flow quality is a function of the environmental lighting and texture~\cite{wannenwetsch2017probflow}, ego-motion performance is also dependent upon these factors.

Typically, image flow likelihood models assume \emph{a priori} an identically distributed Gaussian noise model.
This assumption is inseparable from the common SVD-based essential and fundamental matrix fitting routines used in solving the ego-motion problem~\cite{hartleyzisserman:2004}.   
In this paper, we propose a texture-dependent, data-driven likelihood model that is inspired by the aperture problem.
The structure and parametrisation of the likelihood model are chosen \emph{a posteriori} from empirical data.

\subsection{Related work} 
\label{sub:related_work}

Current work in the area of vision likelihood models has been focused on Gaussian likelihood models. 
Wannenwetsch et al.~\cite{wannenwetsch2017probflow} pose joint flow estimation and uncertainty quantification for image flow using variational techniques. 
They assume that image flow error can be modelled as a Gaussian mixture with independent noise in the two principle image axes; however, this leads to poor confidence estimates in non-principle directions. 
Ilg et al.~\cite{ilg2018uncertainty} train multiple instances of FlowNetS, a convolutional neural net approach to estimating image flow described by \cite{dosovitskiy2015flownet}, on independent datasets. 
A mixture of neural nets predicts the flow given an image pair and computes the sample mean and covariance of the FlowNetS instance outputs. 
This leads to an implicit Gaussian assumption on the flow error. 
Sun et al.~\cite{sun2018bayesian} pose an MCMC-Gibbs approach to estimate the optimal parameters of a Horn and Schunck based image flow algorithm and then approximate the posterior distribution of image flow using the sample mean and covariance of the output of the Gibbs sampler, enforcing a Gaussian model on the flow distribution similar to \cite{ilg2018uncertainty}. 

More recently, attention has been paid to the shape and characteristics of the image flow error distribution. 
Kendall and Gal \cite{kendall2017uncertainties} present hypotheses on the types of noise that are characteristic of vision sensors. 
They highlight heteroscedastic uncertainty, where measurement noise is a function of attributes of the sensor data; such as measurement magnitude, or other transformations of the measurements. 
Heteroscedastic uncertainty underpins most of the recent developments in vision likelihood models.
Min et al.~\cite{min2020voldor} present an empirically derived log-logistic likelihood model for image flow which models heteroscedastic uncertainty as a function of the flow vector magnitude.
This method has been dubbed VOLDOR, and has shown promising ego-motion results when evaluated on the KITTI dataset. 
In a similar vein, \cite{yang2020d3vo} use a deep neural net to learn an \emph{a priori} Laplacian distribution on the flow error, where the heteroscedasticity is modelled by the violation of the brightness constancy assumption.

Our approach offers a different perspective for empirically derived, heteroscedastic optical flow likelihood models; 
we present a texture based likelihood model, which is inspired by directions of texture observed in the aperture problem.

\subsection{Problem description} 
\label{sub:problem_description}

To perform maximum likelihood estimation or Bayesian inference with optical flow, we need to evaluate the likelihood function $p(\mathbf{y}_k|\mathbf{x}_k)$ for a given flow measurement $\mathbf{y}_k$ and state $\mathbf{x}_k$.
Unfortunately, this likelihood function is not well understood for a large family of optical-flow-based measurement systems.
Since the measurement uncertainty is dominated by extrinsic environmental texture rather than intrinsic sensor noise, the characterisation of $p(\mathbf{y}_k|\mathbf{x}_k)$ is further complicated.

To make this problem tractable, we seek a parametric likelihood function $q_\theta(\mathbf{y}_k|\mathbf{x}_k)$ 
with parameters $\theta = \theta(t)$ to enable a heteroscedastic dependence on image texture, $t$.
We assume we can generate a sufficiently large number of flow samples $\mathbf{y}_k^{(i)} \sim p(\mathbf{y}_k|\mathbf{x}_k)$ and have access to the corresponding ground truth state $\mathbf{x}_k$ over a wide range of environmental texture for a given two-frame optical flow algorithm. 
Specifically, we aim to find a suitable parametric form for $q_\theta$ for two-frame optical flow and then find the parameter schedule $\theta(t)$ that minimises $\mathcal{D}_\text{KL}(p \Vert q_\theta)$ for each flow algorithm.

\subsection{Contributions} 
\label{sub:contributions}

This paper develops a likelihood model for two-frame optical flow that is suitable for maximum likelihood estimation or Bayesian inference. 
In this study, the following contributions are made: 
(1) a texture-scheduled Laplace Cauchy mixture (LCM) to model image flow distribution,
(2) a data-driven method to calibrate LCM likelihood model; and 
(3) LCMSAC, a variant of RANSAC that employs an LCM inlier model.


\usetikzlibrary{patterns}
\tikzset{%
    aperture/.pic={
    \path [draw=black, rotate=#1-90, pattern=north east lines]
        (-3/4,-3/2) rectangle (3/4,3/2);
    \path [draw=black, rotate=#1-90, scale=1/8]
        (0,14) -- ++(-1,0) -- ++(0,2) -- ++(-1,0) -- ++(2,2)
        -- ++(2,-2) -- ++(-1,0) -- ++(0,-2) -- cycle;
    \path [fill=white, draw=black, even odd rule] 
        circle [radius=2/3] (-1,-1) rectangle (1,1);
    \node (e1) at (-0.8,0.5) {$\vec{e}_1$};
    \draw [->, red, thick, shorten >=-1pt] (0,0) to (-0.7,0.7);  
    \node (e2) at (0.8,0.4) {$\vec{e}_2$};
    \draw [->, blue, thick, shorten >=-1pt] (0,0) to (0.7,0.7);  
    \node (i1) at  (-1.25,2.25) {$\vec{i}_1$};
    \draw [->, thick, shorten >=-1pt] (-2,2) to (-1,2);  
    \node (i2) at (-1.75,1) {$\vec{i}_2$};
    \draw [->, thick, shorten >=-1pt] (-2,2) to (-2,1); 

  }
}

\section{Theory and Approach} 
\label{sec:theory}

We investigate the structure matrix~\cite{harris1988combined} as an indicator of flow quality in two parts.
Firstly, we assume the structure tensor is sufficient to describe the quantities and directions of texture associated with a flow measurement. 
Secondly, we assume the image flow error is directly related to the level of texture associated with the flow measurement. 
Under these assumptions, we formulate our approach to generate an image flow likelihood model. 

\subsection{Notation} 
\label{sub:notation}

We denote the set of basis vectors for the camera as $\{c\}$, world fixed coordinate system as $\{n\}$, image basis as $\{i\}$ and eigenbasis as $\{e\}$. 
We denote position a vector from the camera centre $C$ to some point $P$ expressed in the camera basis $\{c\}$ as $\mathbf{r}_{P/C}^{c} \in \mathbb{R}^3$.
Rotation matrices used to express vectors in different bases are denoted $\mathbf{R}^{\text{to}}_{\text{from}} \in \mathsf{SO}(n)$.
For example, a vector expressed in the camera basis $\mathbf{r}_{\cdot/\cdot}^{c}$ can be described in the world fixed coordinate system as $\mathbf{r}_{\cdot/\cdot}^{n} = \mathbf{R}^{n}_{c}\mathbf{r}_{\cdot/\cdot}^{c}$.
Homogeneous transformations also follow this notation, where $\mathbf{T}^{n}_{c}  = \left[\begin{smallmatrix}\mathbf{R}^{n}_{c} & \mathbf{r}_{C/N}^{n}\\\mathbf{0}^\mathsf{T} & 1 \end{smallmatrix}\right] \in \mathsf{SE}(3)$.
For image intensities, we express them at frame $k$ by $\mathcal{I}_{k}$.

\subsection{Two view geometry} 
\label{sub:background}

We assume the camera model can be fully described by the kinematic relationship between image pixels and their associated direction vectors.
Let $\mathtt{p2v}\colon \mathbb{R}^2 \rightarrow \mathsf{S}^2$ be the pixel to unit vector mapping and let $\mathtt{v2p}\colon \mathsf{S}^2 \rightarrow \mathbb{R}^2$ be the unit vector to pixel mapping. 
We write the projection of a pixel $j$ from frame $k-1$ in frame $k$ using the following relationship:
\begin{multline}
    \mathbf{g}^{(j)}(\mathbf{x}_k; \mathbf{x}_{k-1}, \mathbf{p}_{k-1}^{(j)}, \mathbf{M}) = \\
    \mathtt{v2p}\left( \mathbf{C}\left(\mathbf{T}^{k}_{k-1}
    \begin{bmatrix}
    \mathtt{p2v}(\mathbf{p}^{(j)}_{k-1})\\\rho(\mathbf{p}^{(j)}_{k-1}, \mathbf{x}_{k-1}, \mathbf{M})
    \end{bmatrix}
    \right)\right), \label{eq:predicted_pix}
\end{multline}
where $\mathbf{C} = \bmat{ \mathbf{I}_{3\times 3} & \mathbf{0}_{3 \times 1} }$, 
$
    \mathbf{T}^{k}_{k-1}  = \big(\mathbf{T}^{n}_{c}(\mathbf{x}_{k})\big)^{-1}\mathbf{T}^{n}_{c}(\mathbf{x}_{k-1})
$,
\begin{equation}
    \mathbf{T}^{n}_{c}(\mathbf{x})  = 
    \begin{bmatrix}
        \mathbf{R}(\boldsymbol{\Theta}^{n}_{c}) & \mathbf{r}_{C/N}^{n}\\\mathbf{0}^{\mathsf{T} } & 1
    \end{bmatrix} \in \mathsf{SE}(3)
\end{equation}
is the homogeneous transform between the camera and the world and
$
\mathbf{x} =
\left[
    \begin{smallmatrix}
        \mathbf{r}_{C/N}^{n} \\
        \boldsymbol{\Theta}^{n}_{c}
    \end{smallmatrix}
\right]
$
is the state, where $\boldsymbol{\Theta}^{n}_{c}$ are the Euler angles. 
The function $\rho(\mathbf{p}^{(j)}_{k-1}, \mathbf{x}_{k-1}, \mathbf{M})$ is the inverse depth map associated with a pixel $\mathbf{p}^{(j)}_{k-1}$, state $\mathbf{x}_{k-1}$ and map $\mathbf{M}$.

Using \eqref{eq:predicted_pix}, image flow can be prescribed in the image basis, by the camera pose as, 
\begin{equation}
    \label{eq:hx:short}
    \mathbf{h}^{(j)}(\mathbf{x}_{k}) = \\
    \mathbf{g}^{(j)}(\mathbf{x}_k) - \mathbf{p}^{(j)}_{k-1},
\end{equation}
where $\mathbf{h}^{(j)}(\mathbf{x}_{k}) \in \mathbb{R}^{2}$ and we omit the known parameters for brevity.
Therefore, image flow can be expressed by the translation of a pixel through the geometric relationship between the environmental structure and the pose of the camera $\mathbf{x}_{k}$. 
We use this geoemetric relationship to evaluate ground-truth flow in a static scene for constructing our proposed likelihood model.

\subsection{Structure tensor and eigenbasis} 
\label{sub:structure_matrix_and_eigenbasis}

\begin{figure}[h!]
    \centering
    \begin{subfigure}{.3\columnwidth}
        \centering
        \begin{tikzpicture}
            \path (0,3) node {} (-3,3) pic {aperture={90}};
        \end{tikzpicture} 
        \caption{Vertical motion}
        \label{fig:aperture_90}
    \end{subfigure}
    \hspace*{1cm}
    \begin{subfigure}{.3\columnwidth}
        \centering
        \begin{tikzpicture}
            \path (0,3) node {} (-3,3) pic {aperture={135}};
        \end{tikzpicture} 
        \caption{Diagonal motion}
        \label{fig:aperture_135}
    \end{subfigure}
    \caption{
    Pathological case for ambiguous motion of \subref{fig:aperture_90} and \subref{fig:aperture_135} when viewed through an aperture. 
    }
    \label{fig:structure_matrix}
\end{figure}
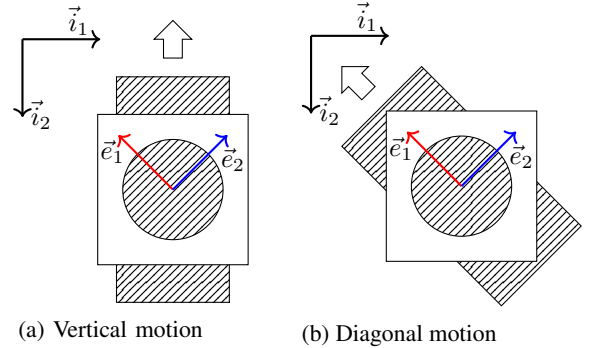

The motion of an edge within a small window of an image may be ambiguous due to the aperture effect.
In Fig.~\ref{fig:structure_matrix}, the global motion perpendicular to the gradient of the image intensity can be readily determined; however, global motion parallel to this gradient is ambiguous. 
Thus, the apparent motion as observed through this aperture will always be in the direction perpendicular to the edges contained in the window~\cite{shi1994}.

The structure tensor \cite{harris1988combined} summarises the distribution of the image intensity gradient over a window. 
Since the structure tensor is symmetric positive semi-definite, its eigenvalues are non-negative and its eigenvectors form an orthonormal basis. 
This matrix is important for feature tracking and optical flow algorithms, which typically use the eigenvalues as an indicator of feature quality. 
The following characterisations are typically proposed: (i) flat response: both eigenvalues are low; (ii) edge response: a single eigenvalue is considerably larger than the other; and (iii) corner response: both eigenvalues are sufficiently high.
If both eigenvalues of a candidate feature are above a certain threshold, it is considered a good point feature to track \cite{shi1994}.

In contrast, we assume there is a continuum of flow quality across the texture space of an image, we exploit the eigendecomposition of the structure tensor to investigate the relation between texture and flow error.

The structure matrix can be factored as
\begin{align}
    \mathbf{S} & = \mathbf{R}^{i}_{e} \boldsymbol\Lambda (\mathbf{R}^{i}_{e})^{\mathsf{T}}, 
\end{align}
where $\mathbf{R}^{i}_{e} \in \mathsf{SO}(2)$ is a matrix with the eigenvectors of $\mathbf{S}$ as its columns, which can be interpreted as the rotation matrix which transforms a vector in the eigenbasis $\{e\}$ to the image basis $\{i\}$, and $\boldsymbol\Lambda = \operatorname{diag}(t_1,t_2)$
is a diagonal matrix of the corresponding eigenvalues.
The eigenbasis is shown in Fig.~\ref{fig:structure_matrix} and the association with directions of texture indicated.
Flow is observable along $\vec{e}_1$, the direction associated with the dominant eigenvalue, but not along $\vec{e}_2$.
Therefore, the eigenbasis of the structure tensor reveals directions in which we can make claims about observed motion. 
Additionally, the diagonalisable nature of the structure matrix infers independence or decoupling of flow measurements in the eigenbasis---a useful property for comparing texture levels to flow error.

As typical with two-image flow algorithms, we form the structure matrix $\mathbf{S}$ for each pixel from the image intensity $\mathcal{I}_{k-1}$. 
Since this matrix is used to inform the estimated image flow for the current time index ${k}$, we express the image flow in the eigenbasis of the structure matrix as follows:
\begin{align}
    \mathbf{y}^{e}_{k} = \mathbf{R}^{e}_{i} \mathbf{y}^{i}_{k},
\end{align}
where $\mathbf{y}^{i}_{k}\in \mathbb{R}^2$ is the image flow expressed in the image basis, $\mathbf{R}^{e}_{i} = (\mathbf{R}^{i}_{e})^\mathsf{T}$ and $\mathbf{y}^{e}_{k} $ is the flow expressed in the eigenbasis.

\subsection{Assumptions} 
\label{sub:assumptions}

We partition the flow vectors into their components expressed in the eigenbasis and assume these components are independent.
This enables edge information to be used for image flow in the direction of its associated eigenvector, which is typically ignored~\cite{shi1994}. 
We will drop the $k$ indexing for brevity.

Let $
\mathbf{Y}^{i}=
\left[
\mathbf{y}^{i, (1)}, \mathbf{y}^{i, (2)}, \dotsc, \mathbf{y}^{i, (n)}
\right]
\in \mathbb{R}^{2 \times n}
$ denote the flow vector field in the image basis, and let
$
    \mathbf{Y}^{e} 
    =\left[
        \mathbf{y}^{e, (1)},\mathbf{y}^{e, (2)},\dotsc, \mathbf{y}^{e, (n)}
    \right]
    \in \mathbb{R}^{2 \times n}
$
denote the flow vector field in the eigenbasis, where
$\mathbf{y}^{e, (j)} = \mathbf{R}^{e, (j)}_{i}\mathbf{y}^{i, (j)},\, \forall j$.
As with~\cite{hartleyzisserman:2004}, we assume independence of image flow vectors, i.e.,
$ 
q_\theta(\mathbf{Y}^{e}|\mathbf{x}) = \prod_{j=1}^n q_\theta(\mathbf{y}^{e, (j)}|\mathbf{x})
$.

The eigenbasis independence assumption allows us to make use of the eigenbasis flow components 
$\mathbf{y}^{e, (j)}=[
    y^{e, (j)}_{1}, 
    y^{e, (j)}_{2}
]^{\mathsf{T} }$
to write the joint component distribution as 
$q_\theta(\mathbf{y}^{e, (j)}|\mathbf{x}) = 
q^1_\theta(y^{e, (j)}_{1}|\mathbf{x})\,
q^2_\theta(y^{e, (j)}_{2}|\mathbf{x})
$.
This assumption allows us to compare flow vector components according to their individual levels of texture.
Therefore, we assume that the likelihood model can be expressed in the form,
\begin{align}
    q_\theta(\mathbf{Y}^{e}|\mathbf{x})
    = \prod_{j=1}^{n}
        q^1_\theta(y^{e, (j)}_{1}|\mathbf{x})\,
        q^2_\theta(y^{e, (j)}_{2}|\mathbf{x}).
\end{align}
We therefore form proposal distributions of the form $q_\theta(y| \mathbf{x} )$, which models the likelihood of the flow component $y\in \mathbb{R}$ in the eigenbasis.

\subsection{Error distribution} 
\label{sub:error_distributions}

We aim to find a parametric distribution that characterises the flow error in the eigenbasis.
The flow field errors in their respective eigenbases are defined as 
\begin{equation}
    \mathbf{Z}^{e} (\mathbf{x}) = 
    \left[
        \mathbf{z}^{e,(1)}(\mathbf{x}), \mathbf{z}^{e,(2)}(\mathbf{x}), \dotsc, \mathbf{z}^{e,(n)}(\mathbf{x})
    \right]
    \in \mathbb{R}^{2 \times n}, \label{eq:eigenerrorfield}
\end{equation}
where the $j$th pixel flow error vector in the eigenbasis is,
\begin{equation}
    \mathbf{z}^{e,(j)}(\mathbf{x})  = \mathbf{R}^{e,(j)}_{i}(\mathbf{y}^{i,(j)} - \mathbf{h}^{(j)}(\mathbf{x})) , 
\end{equation}
$\mathbf{h}^{(j)}(\mathbf{x})$ is given by \eqref{eq:hx:short} and $\mathbf{x}$ is the camera state.
The eigenbasis flow error components for the $j$th vector are denoted,
$
\mathbf{z}^{e,(j)}(\mathbf{x}) = 
    [
        z^{(j)}_{1},
        z^{(j)}_{2}
    ]
    ^{\mathsf{T} }
$.
The field of texture is denoted
\begin{equation}    
    \mathbf{T} = \left[
        \mathbf{t}^{(1)}, 
        \mathbf{t}^{(2)}, 
        \dotsc, 
        \mathbf{t}^{(n)}
    \right] 
    \in \mathbb{R}^{2\times n},\label{eq:texturefield}
\end{equation}
where the texture components are denoted 
$
\mathbf{t}^{(j)} = 
    [
        t^{(j)}_{1},
        t^{(j)}_{2}
    ]
    ^{\mathsf{T} }
$. 
We denote 
the error set as $\mathcal{Z} = \big\{
z^{(1)}_{1}, z^{(1)}_{2},
z^{(2)}_{1}, z^{(2)}_{2},
\dotsc,
z^{(N)}_{1}, z^{(N)}_{2}
\big\}$
and the texture set as $\mathcal{T} = \big\{
t^{(1)}_{1}, t^{(1)}_{2},
t^{(2)}_{1}, t^{(2)}_{2},
\dotsc,
t^{(N)}_{1}, t^{(N)}_{2}
\big\}
$.
For simplicity we denote a single element of the eigenbasis flow error set as $z_i \in \mathbb{R}$ and its associated element in the texture set as $t_i \in \mathbb{R}$. 

We consider Gaussian, log-logistic and Laplace-Cauchy mixture (LCM) distributions to describe the empirical error distribution. 
The Gaussian distribution is chosen as it is representative of the noise model assumed in the SVD essential and fundamental matrix fitting routines that underpins existing ego-motion methods~\cite{hartleyzisserman:2004}.
The log-logistic distribution is considered as it has success in describing the error distribution when scheduled by flow magnitude \cite{min2020voldor}.
The LCM distribution is chosen as it is congruent with the empirical error distribution that is discussed in Sec.~\ref{sec:discussion}.
The LCM is defined as 
\begin{align}
    \begin{split}
        \mathcal{LCM}(x;\theta)
        &=  \dfrac{1}{2}w_L\tan\left({\tfrac{\pi}{2}\beta}\right) \exp\Big( {-|x|\tan\left({\tfrac{\pi}{2}\beta}\right)}\Big) \\
        & \qquad + (1 - w_L)\dfrac{\gamma}{\pi(\gamma^2 + x^2)},
    \end{split}
\end{align}
where $0 < \beta < 1$ is the normalised angle of the Laplace log-space component gradient, $\gamma > 0$ is the scaling parameter of the Cauchy component and $0 \leq w_L \leq 1$ is the weight of the Laplace component and $\theta = \bmat{\beta & \gamma & w_L}$.

\subsection{Fitting error distributions} 
\label{sub:fitting_error_distributions}
Since the true distribution of flow error $p(z_i| t_i)$ is unknown, we form a proposal distribution $q(z_i | t_i, \theta_i)$, where $\theta_i$ denotes the parameters of the proposal distribution for a given texture level.

To capture all texture levels, we represent the parameters of the distribution using a look-up table (LUT) and linear interpolation. 
The LUT $\theta(t; \mathbf{u})$ maps a texture level $t \in \mathbb{R}$ to the distribution parameters $\theta \in \mathbb{R}^{P}$, where $\mathbf{u} \in \mathbb{R}^{P\times M}$ is the set of $M$ LUT entries.
Therefore, we model our conditional distribution as $q(z\, |\, \theta(t; \mathbf{u}))$.

The LUT entries $\mathbf{u}$ are found by minimising the Kullback Liebler (KL) divergence between the proposal distribution and the empirical data distribution,
\begin{align}
    \mathcal{D}_\text{KL}(p || q_\theta) & = \int p(z|t)\log\dfrac{p(z|t)}{q(z;\,\theta(t; \mathbf{u}))}\,\mathrm{d}z\nonumber\\
    & = \text{const.} - \int p(z|t) \log q(z;\,\theta(t; \mathbf{u}))\,\mathrm{d}z. 
    \label{eq:kl_divergence}
\end{align}
The non-constant term is the expectation of the log of our proposal distribution over the domain of $e$. 
While we cannot evaluate $p(z|t)$ we can draw samples of our data distribution $z_i \sim p(z|t_i)$.
Therefore, we use the sample expectation over the elements of the training set $(z_i, t_i) \in \mathcal{Z}\times\mathcal{T}$ and minimise the following cost,
\begin{align}
    \mathbf{J}(\mathbf{u}) & = -\dfrac{1}{N}\sum^{N}_{i=1}\log q(z_i;\, \theta(t_i; \mathbf{u})).
    \label{eq:nll_inducing_params}
\end{align}
Given a large number of samples, minimising the KL divergence is then approximately equivalent to solving the following minimisation problem,
\begin{align}
    \mathbf{u}^* = & \arg \min_{\mathbf{u}} \mathbf{J}(\mathbf{u}) \label{eq:min_kl_div}.
\end{align}

The resulting $\mathbf{u}^*$ provides the optimal LUT parameters to describe proposed distribution over the texture space.
Using the afore-mentioned proposal distribution, we model our image flow likelihood function as, 
\begin{align}
  q_\theta(\mathbf{Y}^{e} | \mathbf{x}) = q(\mathbf{Z}^e(\mathbf{x}); \theta(\mathbf{T}; \mathbf{u})),
  \label{eq:proposal:distribution}
\end{align}
where  $\mathbf{Y}^{e}$ is the flow measurement field, $\mathbf{x}$ is the robot state, $\mathbf{Z}^{e}(\mathbf{x})$ is  given by \eqref{eq:eigenerrorfield}, $\mathbf{T}$ is given by \eqref{eq:texturefield} and $\mathbf{u}$ is the LUT entries that describe the proposal distribution parameters across the texture space.

\subsection{LCMSAC} 
\label{sub:lcmsac}

RANSAC\footnote{RANdom Sampling And Consensus} is a popular method for dealing with flow outliers in ego-motion~\cite{hartleyzisserman:2004}.
This approach attempts to fit a pose hypothesis to the largest viable subset of the measurement data and these inliers are used to perform maximum likelihood ego-motion, while the outliers are rejected as erroneous flow measurements.

In RANSAC ego-motion, a Gaussian likelihood model is typically used to test which components of the measurement set are within the support of a chosen confidence region and then to solve the maximum likelihood problem using only the inlier measurement data.
Since maximising a Gaussian likelihood corresponds to a least squares problem, this leads to the common interpretation of RANSAC as a robust least squares estimator.

We propose to replace the Gaussian likelihood used to both choose and fit the inlier data with our proposed heteroscedastic LCM likelihood model.
To distinguish between RANSAC using a Gaussian inlier model and RANSAC using a LCM inlier model, we denote the latter as LCMSAC.

This generalises the traditional RANSAC ego-motion approach to consider heteroscedastic measurements for which the error bound can be found from a chosen confidence region.
As each flow component error is assumed to be independent in the eigenbasis, we look at the cardinality of the measurement \emph{component} set to validate the fit. 
Inliers are classified as any measurement component that is contained within a chosen confidence region of the LCM distribution. 
The confidence bound is then a function of the distribution parameters over the texture space, which accounts for the heteroscedastic nature of the texture based likelihood model.


\section{Results} 
\label{sec:results}

A series of simulation experiments were performed to fit and validate the proposed likelihood model for both the Lucas Kanade and Farneb\"{a}ck algorithms.
We employed a high-fidelity virtual environment to generate ground truth and estimated flow to compare texture levels to flow error over a sequence of frames.
We present visual odometry ego-motion results to show the performance of the data-driven likelihood model compared to the Gaussian likelihood model, within a RANSAC framework.
Additionally, we apply our approach to the KITTI dataset and obtain competitive  results.

\subsection{Virtual environment and simulation configuration} 
\label{sub:virtual_environment_and_simulation_configuration}

\begin{figure}[t]
    \centering
    \includegraphics[trim=2.8cm 0 0cm 0cm, clip, width=\linewidth]{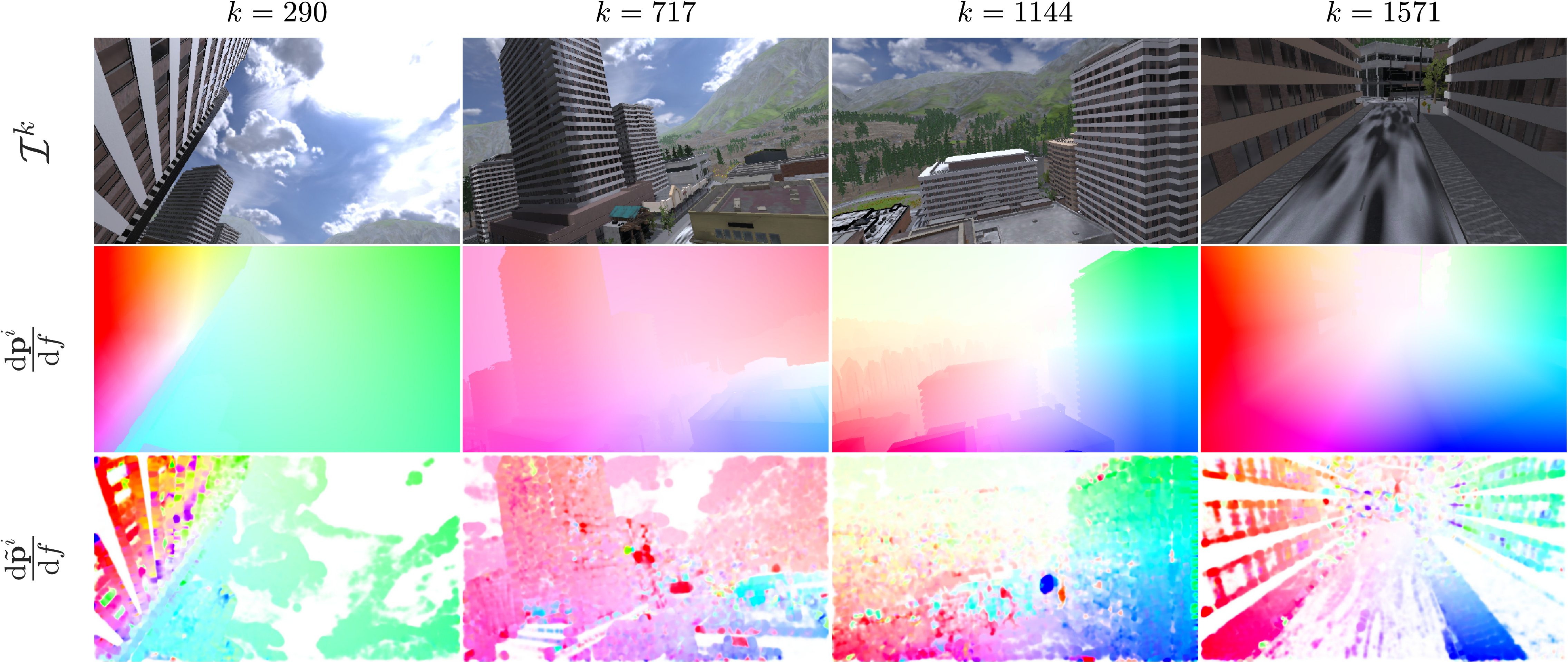}
    \caption{Image flow evaluated using the Farneb\"{a}ck algorithm \cite{farneback2003two} in the Unity3D environment provided by \cite{airsim2017fsr}. 
    Top: Image at the $k$th frame, 
    Middle: hue-encoded ground truth flow,
    Bottom: hue-encoded estimated image flow using the Farneb\"{a}ck algorithm.}
    \label{fig:fb_flowseq}
\end{figure}

We employed the simulation environment provided by \cite{airsim2017fsr} in the Unity3D game engine. 
A virtual camera with a $120^\circ$ field of view and frame dimension $640 \times 360$ operating at 60 frames-per-second is used for simulation experiments.
The environment, camera path, and ground truth and estimated flow measurements are shown in Fig.~\ref{fig:fb_flowseq}. 

\subsection{Learning the model} 
\label{sub:learning_the_correlation}

To learn the relation between texture and flow error in the eigenbasis, we performed experiments using the simulation environment for each flow algorithm. 
The configuration and number of frames used for training each model are given in Table~\ref{tab:correlation_setup}. 

\begin{figure}[h!] 
    \centering
    \includegraphics[
    trim=0cm 0cm 0cm 11.5cm, 
    clip,
    width=\linewidth]{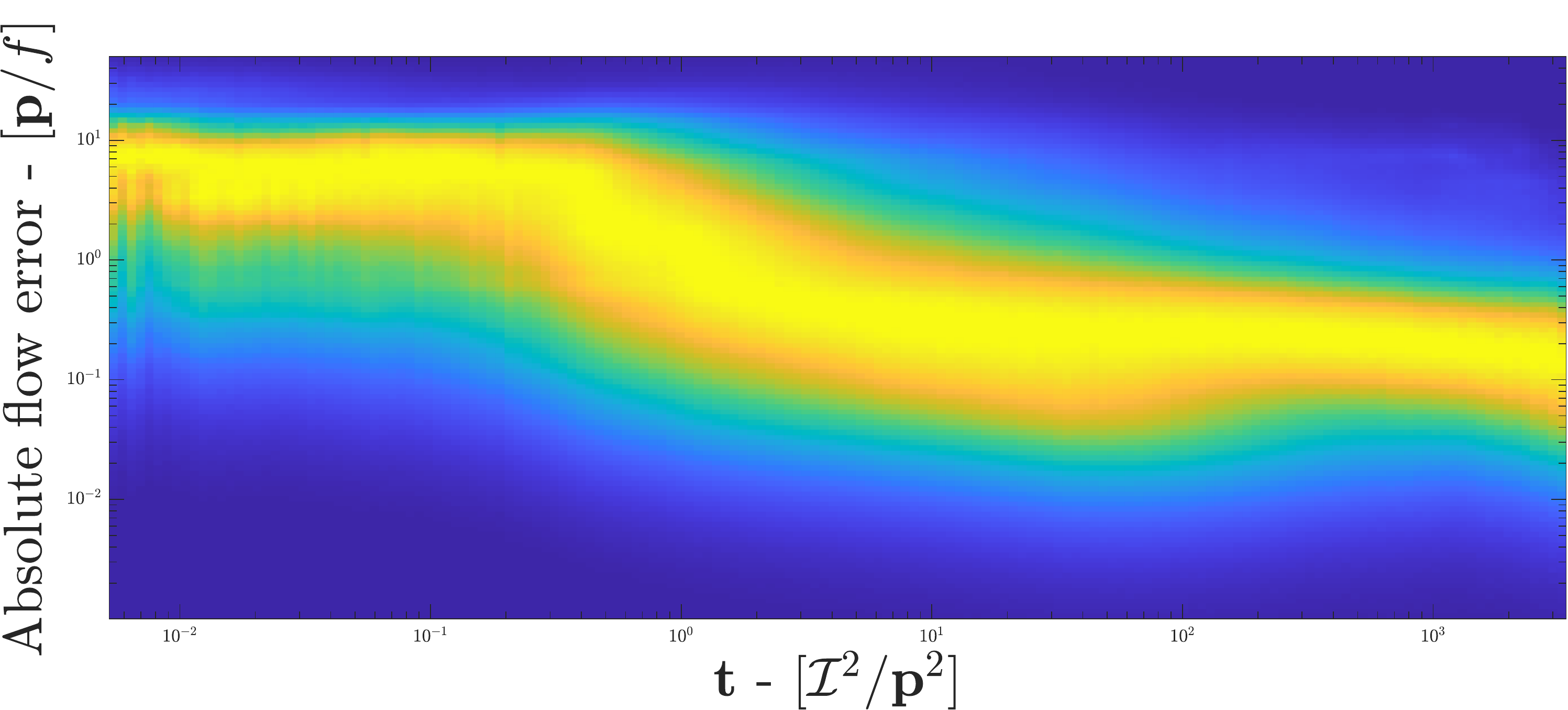}
    \caption{
        Absolute eigenbasis image flow error, in log-scale, for the Farneb\"{a}ck algorithm plotted over the texture range for a simulation sequence with 742 million flow error and eigenvalue components.
    }
    \label{fig:heatmap}
    
\end{figure}

\begin{table}[h!]
	\caption{Training set-up}
	\label{tab:correlation_setup}
	\centering

	\begin{tabular}{l|cp{0.5\linewidth}}
	\hline

	\hline
	\textbf{Algorithm} & \textbf{No. frames}& \textbf{Alg. Parameters} \\
	\hline
		Farneb\"{a}ck & 1651 & Pyramid scale: 0.5, Pyramid~levels:~5, Window~size:~5, Iterations:~5, Gaussian~smoothing~neighbourhood:~15, Smoothing~length~scale:~1.5 \\
	\hline
		Lucas Kanade & 6478 & Window~size:~21, Max pyramid level:~3, Max search iter.:~30, Pyramidal flow eps:~0.01\\
	\hline
	\end{tabular}
\end{table}
We computed the error between ground truth and estimated flow for a training set and \emph{binned} the dataset over several texture ranges in the eigenbasis to inspect the relation.
Figure~\ref{fig:heatmap} demonstrates the relation between eigenbasis flow error and the associated eigenvalues of the structure matrix. 
Investigating slices of texture provides insight into the structure of the data distribution in the eigenbasis. 
This empirical distribution is shown in Fig.~\ref{fig:emp_proposal_distribution}.

\subsection{Fitting the distribution} 
\label{sub:fitting_distribution}

\begin{figure}[h!]
	\centering
	\begin{subfigure}{\linewidth}
		\centering
		\includegraphics[width=\linewidth]{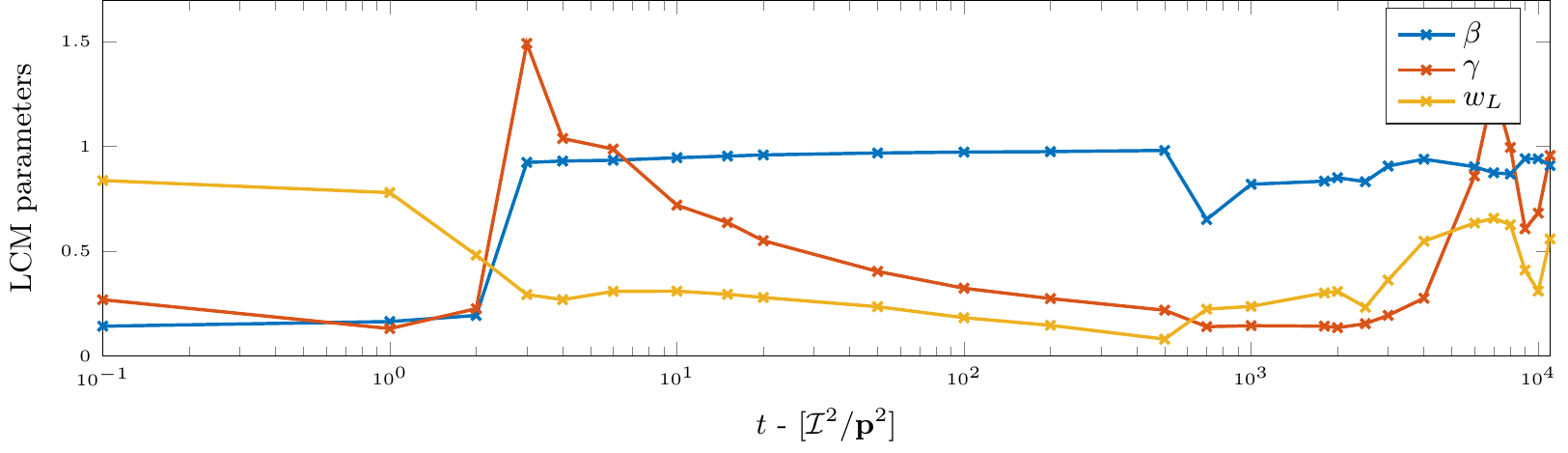}
		\caption{Farneb\"{a}ck}
		\label{fig:lut_fb}
	\end{subfigure}

	\begin{subfigure}{\linewidth}
		\centering
		\includegraphics[width=\linewidth]{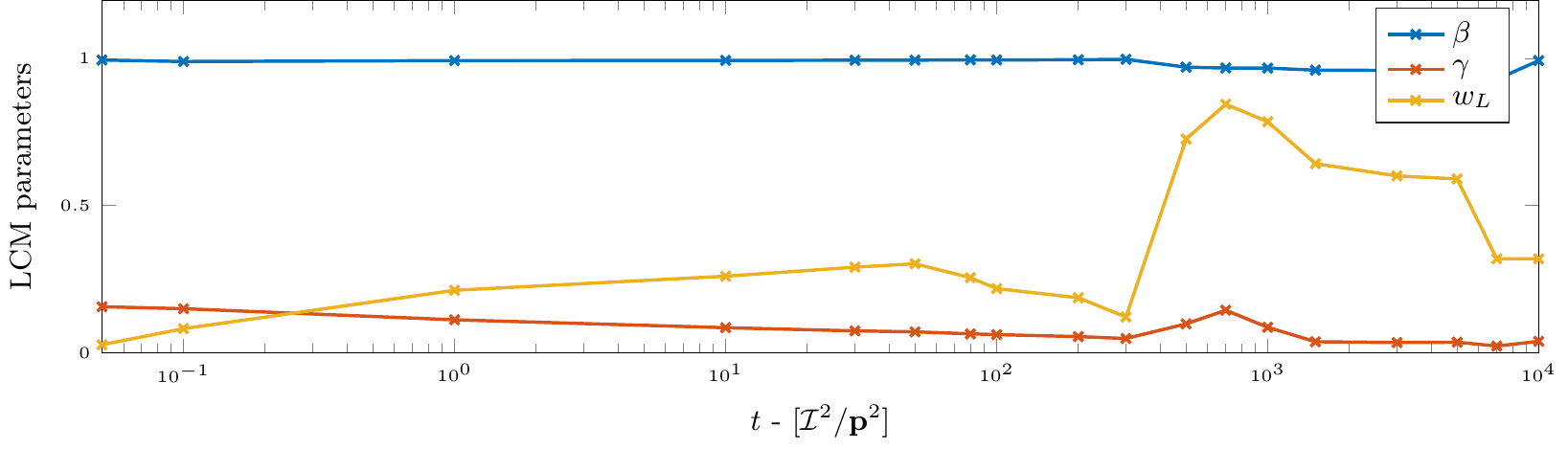}
		\caption{Lucas Kanade}
		\label{fig:lut_lk}
	\end{subfigure}
	\caption{
		Look-up table for distribution parameters over the activated texture space.
	}
	\label{fig:lut_params}
\end{figure}

\begin{figure}[h!]
    \centering
    \begin{subfigure}{\linewidth}
		\centering
		\includegraphics[width=\linewidth]{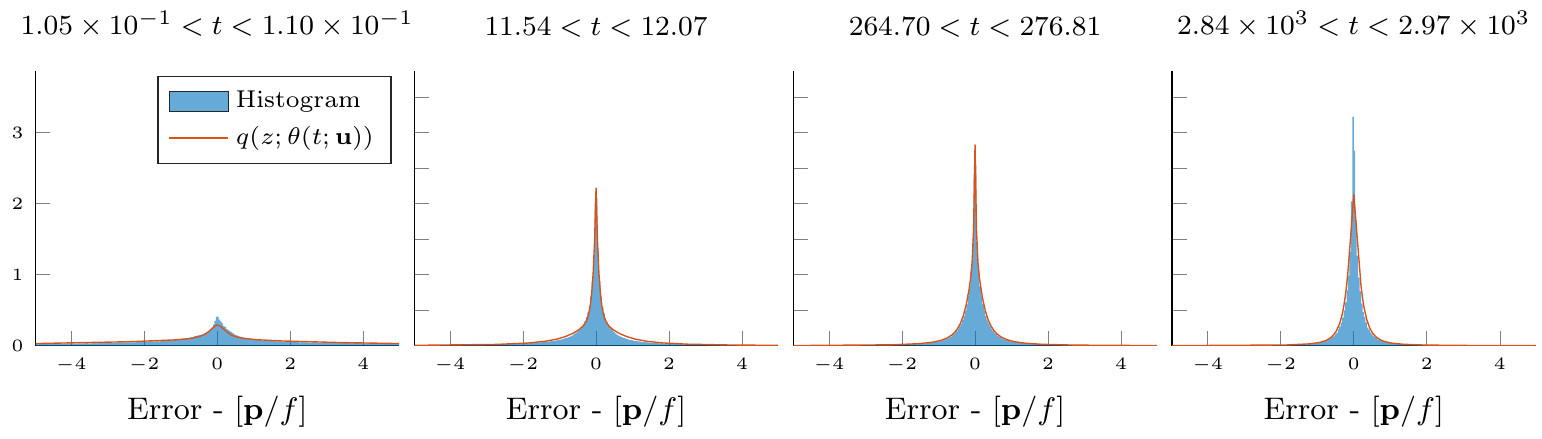}
		\caption{Empirical and LCM distributions (linear scale)}
		\label{fig:lin_fit_dist}
	\end{subfigure}

	\begin{subfigure}{\linewidth}
		\centering
		\includegraphics[width=\linewidth]{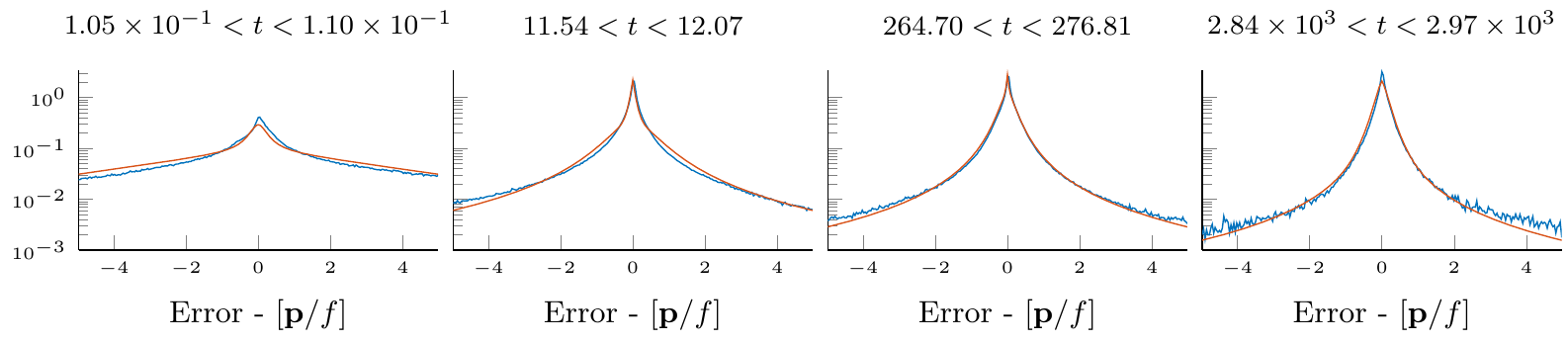}
		\caption{Empirical and LCM distributions (log scale)}
		\label{fig:log_fit_dist}
	\end{subfigure}
	
	\caption{
		Empirical eigenbasis error distribution compared to the fitted error distribution for various ranges of texture for the Farneb\"{a}ck algorithm. 
		}
	\label{fig:emp_proposal_distribution}
    
\end{figure}

Figure~\ref{fig:heatmap} shows an almost monotonic decrease in error with increasing texture and we constructed a look-up table (LUT) to linearly interpolate parameters in the log-texture space.
We placed the LUT entries at texture values shown in Fig.~\ref{fig:lut_params}. 
Using the training datasets of image texture and eigenbasis flow error, we performed regression using \eqref{eq:min_kl_div} to learn the LUT entries for the LCM distribution $q_\theta(x) = \mathcal{LCM}(x; \theta)$ for each flow algorithm. 
The resulting proposal distribution closely agrees with the empirical histogram as shown in Fig.~\ref{fig:emp_proposal_distribution}.

\begin{figure}[h]
	\centering
	\includegraphics[trim=2cm 0 2cm 0, clip, width=\linewidth]{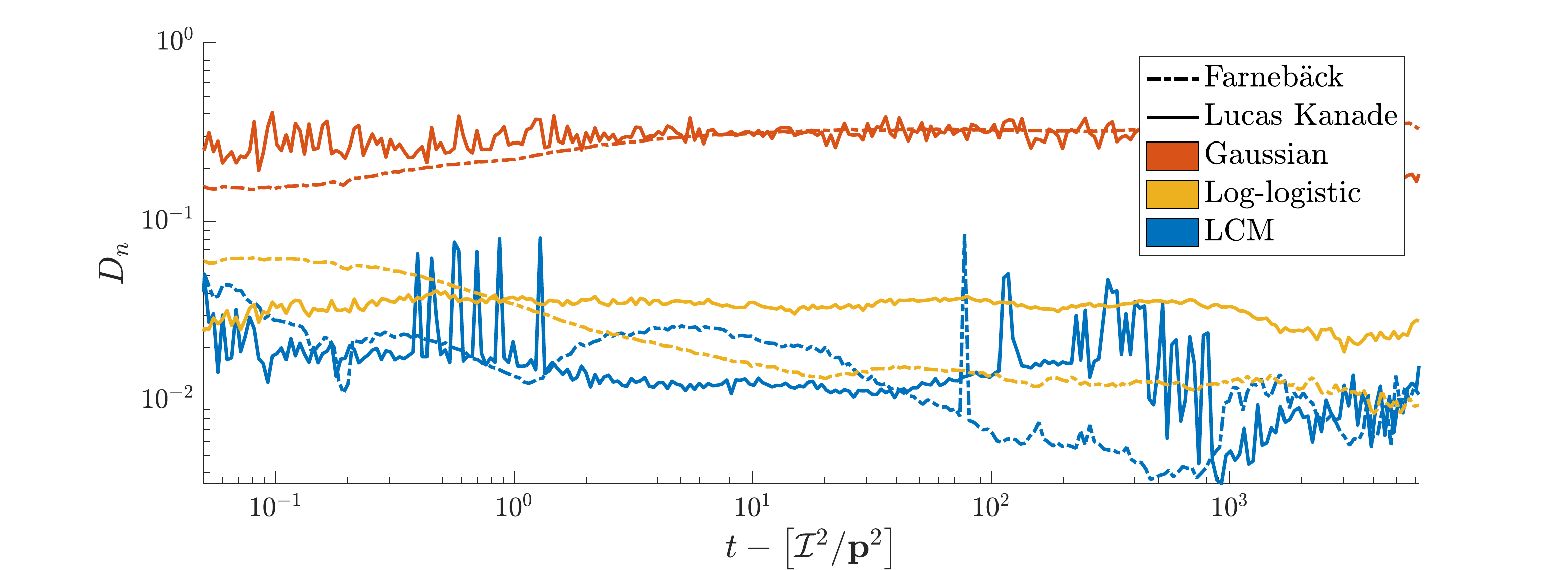}
	\caption{K-S statistic across the texture space for Gaussian, log-logistic and LCM distributions.}
	\label{fig:ks_statistics}
\end{figure}

To quantify the distance between samples from the empirical distribution and our proposed LCM likelihood model, the Kolmogorov-Smirnov statistic is used, which measures the maximum absolute distance between the empirical and proposal CDF,
$D_{n} = \max\limits_{x} |F_{n}(x) - F(x)|$ as shown in \cite{masseyKStest1951}.
Figure~\ref{fig:ks_statistics} shows the K-S statistic for the Gaussian, log-logistic and LCM distributions over the texture space of the simulation data to quantify the \emph{goodness of fit} of the distributions. 

\subsection{Visual odometry example} 
\label{sub:visual_odometry_example}

To demonstrate the advantage of using an empirically congruent image flow error distribution, we run a visual-odometry ego-motion simulation to compare LCMSAC and RANSAC. 
The ego-motion problem is solved using maximum likelihood estimation, which involves solving the following optimisation problem for every image frame:
\begin{align}
	\mathbf{x}_{k}^{*} = \arg \max_{\mathbf{x}_k} \log q_\theta(\mathbf{Y}^{e}_{k} | \mathbf{x}_k),
\end{align}
where we have re-introduced the frame index $k$.
Parameters $\theta$ are constant across the texture space for the Gaussian likelihood used within RANSAC, but for the LCM likelihood vary as a function of texture $\theta = \theta(t; \mathbf{u})$ according to~\eqref{eq:proposal:distribution}.
The inlier threshold for RANSAC is a flow error of 0.5 pixels per frame. 
For LCMSAC, the threshold is a function of the image texture and the 90\% confidence region of the proposed distribution as outlined in Sec.~\ref{sub:lcmsac}.
The proposed likelihood model was modified for the Farneb\"{a}ck algorithm as the dense flow measurements bias the estimator when there is a large portion of low textured, zero flow measurements. 
A texture-based measurement threshold was applied to ignore measurements with texture levels below 50 $\mathcal{I}^2/\mathbf{p}^2$, to ameliorate the bias.
This is similar to the texture thresholding present in the Lucas Kanade algorithm. 

\begin{figure*}[h!]
	\centering
 
	\subcaptionbox{\emph{straight} trajectory\label{fig:ego_straight}}%
		[.49\textwidth][c]%
		{%
    		\includegraphics[trim=2cm 0 2cm 0, clip, width=.48\textwidth]{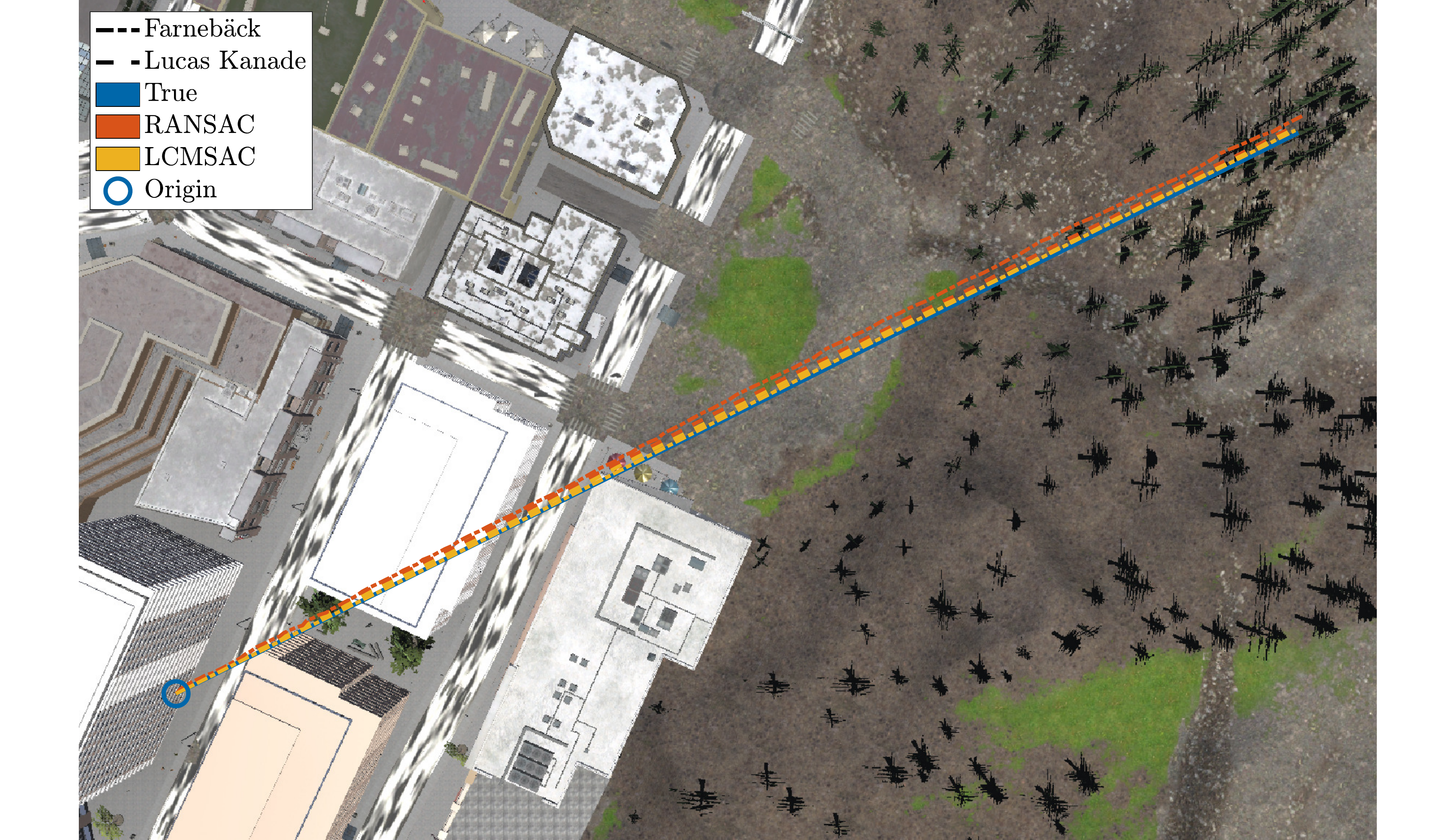}
		}
	\subcaptionbox{\emph{forest-fig8} trajectory\label{fig:ego_fig8}}%
		[.49\textwidth][c]%
		{%
    		\includegraphics[trim=2cm 0 2cm 0, clip, width=.48\textwidth]{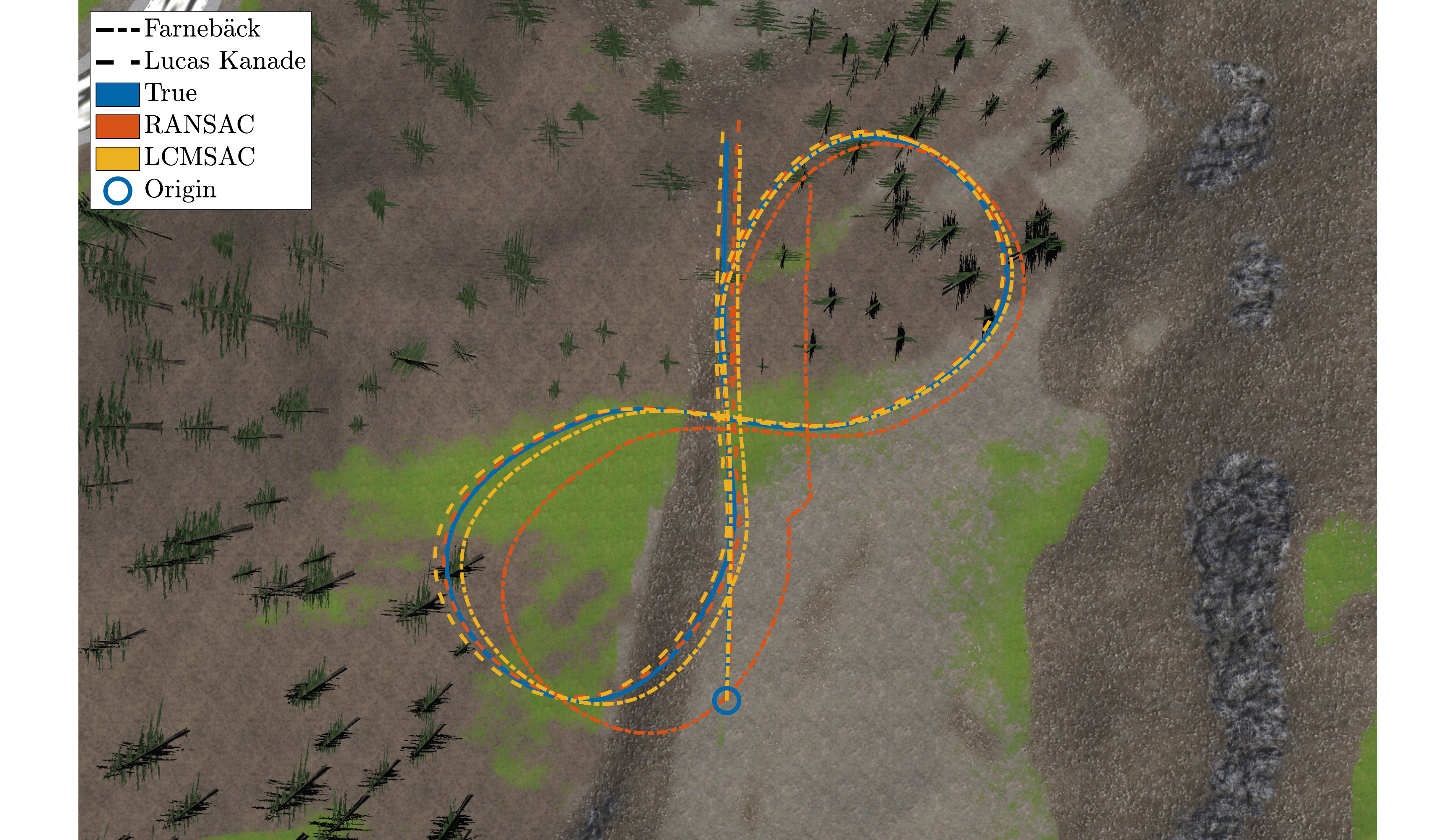}
		}

    \caption{
    Evaluating ego-motion estimation in simulation on both the Farneb\"{a}ck and Lucas Kanade algorithms for the \emph{straight} and \emph{forest-fig8} trajectories. 
    }
    \label{fig:ego_all}
\end{figure*}

RANSAC and LCMSAC estimators are evaluated in the Unity3D environment. 
The estimators are provided with the ground truth depth map so that we can study the effect of the proposed likelihood model directly. 
The disparities between both estimators and ground truth motion are shown in Fig.~\ref{fig:ego_all}.

We compare the estimator drift rates over the same trajectories to evaluate the relative performance of the ego-motion estimators. 
The translational drift rate is given as a percentage of the norm of the translational estimation error divided by the trajectory length.

Table~\ref{tab:ego_results} lists the ego-motion loop closure drift rates for both estimators on the evaluated trajectories. 

\begin{table}[h!]
	\caption{Simulation ego-motion results}
	\label{tab:ego_results}
	\centering

	\begin{tabular}{l|ccc}
	\hline

	\hline
	\textbf{Flow Alg.} & \textbf{Estimator} & \textbf{Eval. Traj.} & \textbf{Drift} $[\%]$ \\
	\hline
		Farneb\"ack 
		 & RANSAC & straight  & 1.410 \\ 
		 & LCMSAC & straight & 0.707  \\
		\cline{2-4}
		 & RANSAC & forest-fig8 & 4.161 \\ 
		 & LCMSAC & forest-fig8 & 0.669\\
		 \hline
		Lucas Kanade 
		
		 & RANSAC & straight  & 0.720  \\ 
		 & LCMSAC & straight & 0.503  \\
		\cline{2-4}
		 & RANSAC & forest-fig8 & 1.513  \\ 
		 & LCMSAC & forest-fig8 & 1.010  \\
	\hline
	\hline
	\end{tabular}
\end{table}

\begin{figure}[h!]
	\centering
	%
	\subcaptionbox{KITTI 01 Sequence\label{fig:ego_lcmsac_kitti_01}}%
		[.4\textwidth][c]%
		{%
    		\includegraphics[
    		trim=2cm 0cm 2cm 1cm, 
    		clip, 
    		width=.4\textwidth]
    		{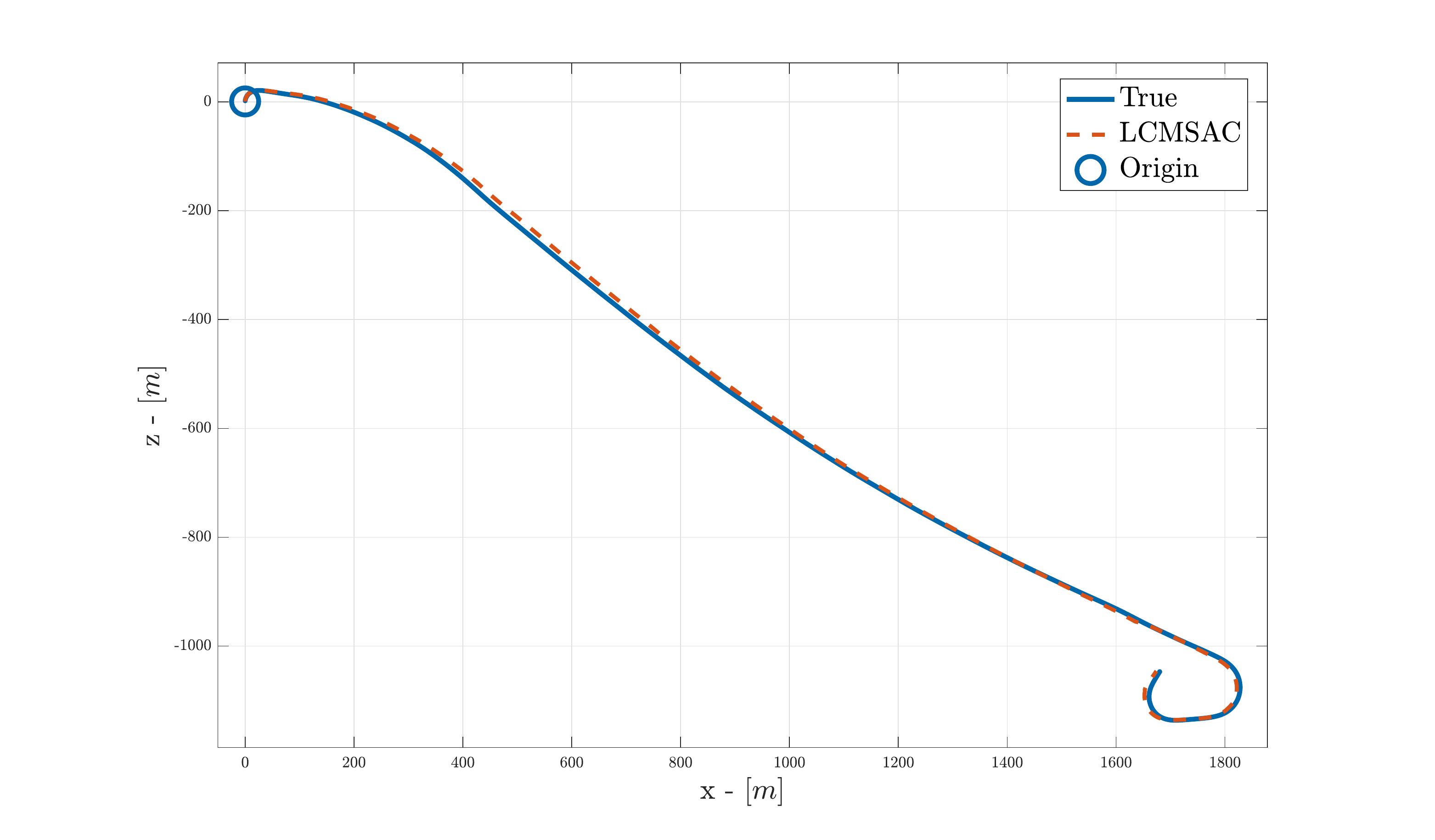}
		}
	\subcaptionbox{KITTI 03 Sequence \label{fig:ego_lcmsac_kitti_03}}%
		[.5\textwidth][c]%
		{%
    		\includegraphics[trim=2cm 2cm 2cm 2cm, clip, width=.4\textwidth]{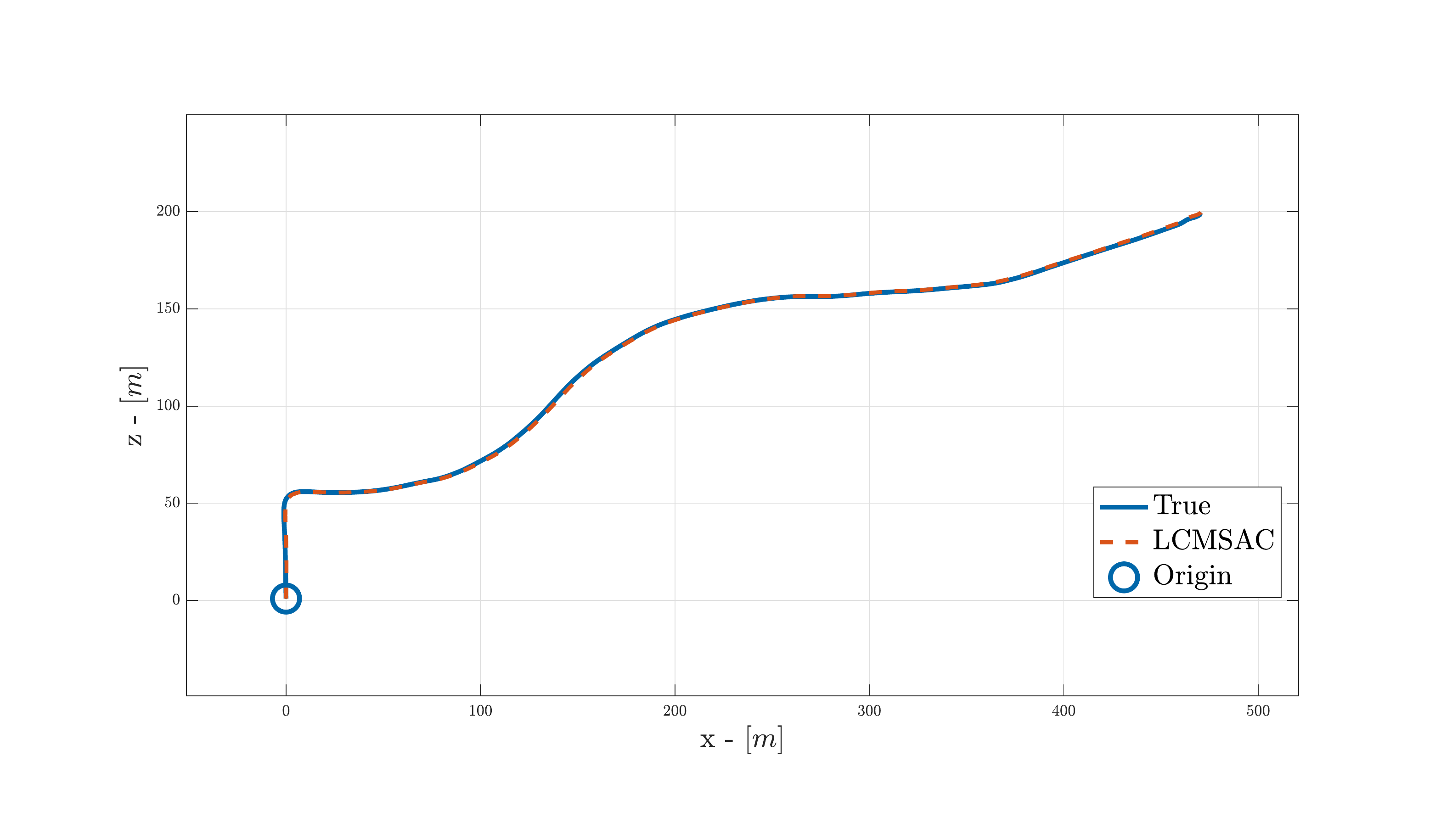}
		}
	\caption{Visual odometry using the proposed likelihood model on the KITTI dataset.}
	\label{fig:kitti_egomotion}
\end{figure}

We use LCM likelihood models, trained in simulation, and apply them to the KITTI dataset to evaluate odometry drift on real-world data. 
The epipolar constraint is exploited to remove the need for depth information in ego-motion by using flow components that are orthogonal to the epipolar line as shown in \cite{williams2010combining}.
Since motion scale cannot be inferred through a single camera, it is recovered using the norm of the GPS measurements between each frame pair---similar to exploiting knowledge of camera height knowledge in \cite{min2020voldor}.
The moving vehicles in the KITTI dataset are rejected as outliers in the LCMSAC algorithm as they violate the static environment assumption.
Results are shown in Fig.~\ref{fig:kitti_egomotion} for the LCMSAC algorithm on a selection of the KITTI trajectories.
Table~\ref{tab:kitti_results} compares KITTI ego-motion results to VOLDOR \cite{min2020voldor}.

\begin{table}[h!]
	\caption{Results on KITTI training sequences 00, 01 and 03. The translational error is averaged over all sub sequences between 100 meters and 800 meters with 100 meter steps.}
	\label{tab:kitti_results}
	\centering

	\begin{tabular}{c|c|c}
	\hline

	\textbf{KITTI Seq.} & \textbf{VOLDOR Drift} $[\%]$ & \textbf{LCMSAC Drift} $[\%]$ \\
	\hline
	00 & 1.09 & \textbf{0.655}\\
	01 & 2.31 & \textbf{0.942}\\
	03 & 1.46 & \textbf{0.536}\\
		
	\hline
	\end{tabular}
\end{table}

\section{Discussion} 
\label{sec:discussion}

Descriptive likelihood models are fundamental for performing maximum likelihood estimation or Bayesian inference. 
Until recently, there has been an absence of empirically validated likelihood models for image flow.
In this paper, we pose a data-driven likelihood function to model the uncertainty of two-frame optical flow and apply it to the 
Lucas Kanade and Farneb\"{a}ck flow algorithms. 

The empirical distributions shown in Fig.~\ref{fig:emp_proposal_distribution} reveal a sharp peak and very wide tails, for which we propose the LCM likelihood with parameters scheduled over the texture space using a LUT.
The combination of the LCM structure and the LUT enables our proposed likelihood model to account for both the characteristics of the empirical distribution and the varying quality of flow for the space of texture.

Likelihood models were trained on a specific training trajectory, and ego-motion was evaluated on alternative trajectories to ameliorate over-fitting and produce a model for general environmental application.
The results, shown in Table~\ref{tab:ego_results}, outline the performance increase by using the LCM inlier model over a Gaussian inlier model within a RANSAC framework. 
LCMSAC reduces the ego-motion drift rate by $49\%$ to $83\%$ with the Farneb\"{a}ck algorithm and by $30\%$ to $33\%$ with the Lucas Kanade algorithm.
The discrepancy of the performance enhancement is primarily due to the texture thresholding native to common Lucas Kanade methods. 
Nevertheless, the results demonstrate a decrease of estimation drift rate of at least $30\%$ using the proposed likelihood model.
These performance increases highlight the sensitivity of the estimator to the selection of likelihood model. 

We demonstrate the generalisation of our method to unseen data by applying simulation trained likelihood models to ego-motion on the KITTI dataset.
The performance of LCMSAC with Lucas Kanade demonstrates the effectiveness of the likelihood model to describe ego-motion---with results that have at least 39\% reduction in translational drift compared to VOLDOR \cite{min2020voldor} on the same sequences.

Although the intent of this paper is to present a texture based likelihood model for optical flow, it is interesting that applying the likelihood model to widely accessible optical flow algorithms and a RANSAC based approach results in ego-motion performance which is comparable with state-of-the-art methods. 
Our ego-motion results, arise from the congruency of our characterised likelihood model with empirical data.
This performance highlights the utility of our proposed likelihood model class for visual odometry, but also the advantage and wider application in enabling optical flow to be used as a sensor within Bayesian estimators.

\subsection{Limitations} 
\label{sub:limitations}

As the proposed likelihood model uncertainty is derived from texture, there are scenarios where the proposed likelihood model may not be suitable.
For example, in the presence of self-similar patterns such as a regular grid, the likelihood model will incorrectly proclaim image flow measurements to be highly trustworthy, without accounting for the possibility of feature mismatch.
This mismatching scenario often occurs in flow algorithms such as Lucas Kanade, but is mitigated by the rejection strategies in sampling-and-consensus methods.

Furthermore, our approach does not assume any uncertainty in the orientation of the eigenbasis.
This may reduce the effectiveness of the proposed likelihood in situations where edge directions are ambiguous.

Our method has been applied to Farneb\"ack and Lucas Kanade algorithms as they are readily available and widely implemented on embedded systems.
Our approach has not yet been applied to any learning-based or variational flow algorithms; however, future work could consider applications to algorithms such as FlowNet, SPyNet, PWC-Net and Horn-Schunck.

\subsection{Applications} 
\label{sub:applications}

The proposed likelihood model is well-suited to the fields of robotics and autonomous systems and can be applied to navigation problems in ground, maritime and aerospace domains.
Recent cases of GPS misdirection and spoofing attacks~\cite{c4ads2017:spoof} highlight the need to fuse additional sensors to enable the detection of attacks and the ability to navigate in GPS-denied environments. 
Vision provides a plethora of information on system ego-motion; however, until recently, empirically derived and validated likelihood models have been scarce.
The inadequacy of traditional likelihood models has halted coherent Bayesian fusion of vision with other sensors.
Our proposed likelihood model helps alleviate this deficiency and enables optical flow to be included within sensor fusion applications.

\section{Conclusion} 
\label{sec:conclusion}
In this paper, we have investigated the structure of the empirical image flow error distribution.
Inspired by the aperture problem, we have presented an empirically derived likelihood model for image flow in the eigenbasis of the structure matrix. 
The proposed likelihood model accounts for both the characteristics of the image flow error distribution and the heteroscedastic uncertainty of image flow over the texture space. 
The empirically derived likelihood model advances the development of essential apparatus for the incorporation of vision within Bayesian sensor fusion; enabling fusion of vision, GPS, LiDAR and IMU data to strengthen trusted autonomy.

\bibliographystyle{IEEEtran}
\bibliography{bibliography}

\end{document}